\documentclass[10pt,runningheads]{llncs}

\PassOptionsToPackage{final}{showkeys}

\usepackage{mystyle}
\usepackage{tpic}
\usetikzlibrary{backgrounds,arrows}
\usepackage{pgfplots}
\usepackage{pgfplotstable}
\usepackage{tikz}
\usepackage{booktabs} % Provides commands, like \toprule, to create high quality ta­bles

\usepackage{amssymb}
\usepackage{latexsym}
\usepackage{url}
\usepackage{xcolor}
\usepackage{graphicx}
\usepackage{subfigure}
\usepackage{amsmath}

\usetikzlibrary{plotmarks}
\usetikzlibrary{arrows.meta}
\usepackage{grffile}
\usepackage{import}
\usetikzlibrary{arrows, shapes, chains}
\usepackage{todonotes}

\usepackage{algorithm}
\usepackage{algorithmic}

\usepackage{paralist}

\newcommand{\rev}[1]{\textcolor{black}{#1}}
\newcommand{\revv}[1]{\textcolor{black}{#1}}
\newcommand{\revvv}[1]{\textcolor{black}{#1}}

\begin{document}

\title{Is Perfect Filtering Enough Leading to Perfect Phase Correction for dMRI data?}
% Does Perfect Filtering Contributs to Perfect Phase Correction?
\titlerunning{Is Perfect Filtering Enough?}

 % For anonymous submission
%\authorrunning{Anonymous\\Anonymous\\Anonymous}
%\author{Anonymous}
\institute{Paper \#471}

 \authorrunning{F.~Liu, J.~Yang, X.~He, L.~Zhou, J.~Feng, D.~Shen}   % abbreviated author list (for running head)
%  %%%Department of Artificial Intelligence, Korea University, Seoul, Republic of Korea \\                                     % 7	
 \author{Feihong~Liu\inst{1,2}  \and Junwei~Yang\inst{3,2} \and Xiaowei~He\inst{1,4}  \and Luping~Zhou\inst{5}~\Letter \and Jun~Feng\inst{1,4}~\Letter \and   Dinggang~Shen\inst{2,6}~\Letter}
 \institute{  
 School of Information Science and Technology,       Northwest University,         Xi'an,      China            \and	  % 1
 School of Biomedical Engineering,                   ShanghaiTech University,      Shanghai,   China          	\and      % 2
 Department of Computer Science and Technology,      University of Cambridge,      Cambridge,  United Kingdom   \and      % 3
 State-Province Joint Engineering and Research Center of Advanced Networking and Intelligent Information Services, School of Information Science and Technology, Northwest University, Xi’an, China                                                                             \and      % 4
 School of Electrical and Information Engineering,   University of Sydney,         Sydney,     Australia        \and      % 5 
 Shanghai United Imaging Intelligence Co., Ltd., Shanghai, China                                                     % 6  
 \\
             \email{luping.zhou@sydney.edu.au},~\email{fengjun@nwu.edu.cn},~\email{dgshen@shanghaitech.edu.cn} % Dinggang.Shen@gmail.com
 }

\maketitle

\begin{abstract}
	% reliably measuring such diffusion scalars
\rev{
    Being complex-valued and low in signal-to-noise ratios, magnitude based diffusion MRI is confounded by the noise-floor that falsely elevates signal magnitude and incurs bias to the commonly used diffusion indices, such as fractional anisotropy (FA). 
    To avoid noise-floor, most existing phase correction methods explore improving filters to estimate the noise-free background phase. 
    In this work, after diving into the phase correction procedures, we argue that even a perfect filter is insufficient for phase correction because the correction procedures are incapable of distinguishing sign-symbols of noise, resulting in artifacts (\textit{i.e.}, arbitrary signal loss). 
    With this insight, we generalize the definition of noise-floor to a complex polar coordinate system and propose a calibration procedure that could conveniently distinguish noise sign-symbols. 
    The calibration procedure is conceptually simple and easy to implement without relying on any external technique, while keeping distinctly effective. 
    % which is easy to implement and remedy the conventional procedures
    % Diffusion MRI is inherently complex-valued and low signal-to-noise ratio, with the noise-floor that falsely elevates magnitude and incurs bias to the commonly used diffusion indices, such as fractional anisotropy (FA). 
    % To avoid noise-floor, most existing phase correction methods explore improving filters to estimate the noise-free background phase. 
    % However, in this work, after diving into the phase correction procedures, we argue that even a perfect filter is insufficient for phase correction because its procedures are incapable of distinguishing sign-symbols of noise, resulting in artifacts (\textit{i.e.}, arbitrary signal loss). 
   %  With this insight, we give a generalized definition of noise-floor in a complex polar coordinate system and propose a calibration procedure to conveniently distinguish noise sign-symbols. 
    % The calibration procedure is conceptually simple, easy to implement without relying on any other technique, while distinctly effective. 
 }
Extensive experimental results, including those on both synthetic and real diffusion MRI data, demonstrate that the calibrated procedures successfully  mitigate artifacts in diffusion MR images and FA maps, with improved accuracy on estimating FA in particular. 
% Our work suggests that a perfect filter is necessary but not enough yet in advancing phase correction technique. % filtering capacity should be necessarily taken into consideration
%%% The title may should be adjusted.
%However, phase correction introduces artifacts, blaming for 

% We argue that filtering capability contributes to phase correciton performance. 
% take much attention to carting to  spatial noise variability, 
% However,

%  properly rotate image contents to be negative, hence, the dark hole issue is effectively avoided, and FA can be accurately estimated ultimately.
% The calibrated procedures properly rotate image contents to be negative, hence, the dark hole issue is effectively avoided, and FA can be accurately estimated ultimately. Conventional focus on the noise variability, while neglecting the filtering performance. 
%\blfootnote{
%	\noindent $^{\star}$ This work was supported in part by NIH grants (NS093842, EB022880, EB006733, EB009634, AG041721, MH100217, and AA012388) and a NSFC grant (11671022).	
%}
\end{abstract}

\section{Introduction}
Diffusion-weighted (DW) imaging enables reliably characterizing white-matter microstructure in-vivo. 
Using large $b$-values for more diffusion weighting during imaging, DW signals on one hand could be acquired in high resolution~\cite{tuch2002high}, but on the other hand are inherently susceptible to low signal-to-noise ratio (SNR)~\cite{aja2016statistical}. As DW signal is by nature complex-valued,
low SNR magnitude based image regions, especially for those in absence of signal, are superimposed by the well-known noise-floor which falsely elevates signal magnitude, incurring estimation bias to the commonly used diffusion indices, \textit{e.g.}, fractional anisotropy (FA)~\cite{eichner2015real}.

%Diffusion-weighted (DW) imaging enables reliably characterizing white-matter microstructure in-vivo. 
%Although large $b$-values assure DW signal in high resolution \cite{tuch2002high}, on the other hand, the signals are inherently susceptible to low signal-to-noise ratio (SNR)~\cite{aja2016statistical}. 
%\rev{As for} low SNR magnitude-based \rev{image regions, especially for those in absence of signal, they} are superimposed by the noise-floor which falsely elevates \rev{magnitude}, incurring estimation bias to the commonly used diffusion indices, \textit{e.g.}, fractional anisotropy (FA)~\cite{eichner2015real}.
%% In reconstructing the magnitude with the sum of square (SoS) approach; and SoS flips the negative noise to positive such that the negative noise cannot be eliminated by averaging across neighboring positive noise, especially for those image regions in absence of signal. Eventually, the average smoothing in the postprocessing procedures would 
%%%% , during magnitude reconstruction,
%% even at regions where it is often expected to be reliable, such as the corpus callosum, where the neuronal orientations are highly coherent

To avoid the noise-floor, instead of reconstructing magnitude, a preprocessing technique named phase correction is employed, which complex-rotates the DW signal so that its real part carries the true signals plus Gaussian distributed noise and its imaginary part contains purely noise that will be discarded. 
The complex rotation is guided by the background phase, estimated by smoothing real and imaginary signals, respectively.
An early phase correction approach employs total variation (TV)~\cite{eichner2015real} for signal smoothing, achieving reliable performance in white-matter characterization. That work also pointed out the direction in catering TV for spatially-varying noise caused by the multi-coil parallel imaging approach. 

% A limitation to wide use of phase correction is ascribed to that the employed  filter cannot account for spatially-varying noise derived from multi-coil imaging and cause artifacts.
% d accuracy in estimating the level of spatially-varying noise so that patching cater filters. 
% so that cater filters to spatailly-varying noise.  %smoothing imaginary and real signals respectively to recover the background phase, which is used to

%%% phase correction, which  employs an smoothing filter, complex rotates the true signal from the imaginary part to the real part, instead of reconstructing magnitude.
%is to preserve image contents while preserving those negative nosies.    

The advance of current phase correction technique is reflected by exploring filters accounting for the noise variability.
To improve TV, a great deal of analyses have been implemented to evaluate the function of its regularization parameter (\textit{i.e.}, $\lambda$)~\cite{pizzolato2016noise}. 
Those works pave to proposing weighted TV (wTV) that generates voxel-wise weights to adjust $\lambda$ according to the estimation of  noise-level~\cite{pizzolato2018automatic}. 
As anticipated, wTV effectively eliminates artifacts, however, \rev{the evaluation of its filtering performance was left in those works.}    
After that, multi-kernel filtering was put forward, which is based on bilateral filtering and adapts the range kernel parameter voxel-wisely to spatially-varying noise and mitigates the artifacts with a better filtering performance~\cite{liu2019mkf}.
Following that, an improved filtering approach for noise-level estimation was also proposed~\cite{pizzolato2020adaptive}, however, the cross-sectional comparison of its filtering capacity was still not provided. 
% adaptively regularize the filtering procedures is also proposed to enhance the ability carting to spatially varying noise. the enhance is posed on the spatially apdativity rather than filtering performance.
%but also report the significant loss of filtering perofrmance of wTV
% Early phase correction approaches simply employ rectangular and triangular convolution kernels, which cannot carter to the spatially varying noise.
%%% While TV is employed in phase correction demonstrating promissing correction performance, its performance rely on the choice of regularization parameter; However, there induces significant black holes and artifacts. 

%%% Challenges
%Typically, considering the spatial noise variance, caused by the multi-coil paralle imaing approach, that the noise level of peripheral image region is smaller than the central location, the SNR of central image region is mostly probable smaller than $1$.

% cond, it is interestingly found that the filtering performance cannot be comparable to TV, however achieved improved
% a weighted TV algorithm 
%, it demosntartes the worst filtering performance and introduces the most servere artifacts in intermidiate results. 
Meanwhile, two interesting phenomena have attracted our attention. % ~\cite{liu2020gaussianization}
First, wTV that performs well in removing artifacts in DW images and FA maps %seems to have only 
achieves inferior performance on the intermediate filtering results,  possibly due to the compromise between regularization adaptivity and filtering capacity.
%First, rather than introducing artifacts to DW images and FA maps, wTV introduces artifacts to intermediate results, seemingly due to a compromise between better regularization adaptability and worse filtering capacity.  
Second, in contrast to wTV, Marchenko–Pastur principal component analysis (MPPCA)~\cite{veraart2016denoising}, which allows accurate estimation of noise-level, surprisingly introduces considerable artifacts in the DW images and FA maps. 
Both phenomena suggest that patching filters might just be a half-measure and \revvv{that} there could be some underlying problems behind phase correction procedures, hindering the noise-floor suppression.
% Those results seemingly support that the filtering performance is not important, however, this is counterintuitive. 
% However, even MPPCA~\cite{veraart2016denoising}, renowned as spatial adaptivity, surprisingly introduces severe artifacts~\cite{liu2020gaussianization}. 
%%% The reason %%% surprisingly introduces arbitrary artifacts especially employing an advanced filter (\textit{i.e.}, Marchenko–Pastur principal component analysis), which phenomenon is counterintuitive and inspires us to figure out the problems underlying phase correction procedures. 

% with a belief that perfect filtering leading is the basis of catering to noise variability,
% After understanding the reason why phase correction falsely introduces artifacts, 
To further investigate this problem, in this paper, we employ the complex polar coordinate system, which is a combination of polar coordinate and Cartesian coordinate, to cautiously analyze the procedures of phase correction  and to insightfully generalize the definition of  noise-floor.
What follows, we introduce  a simple but effective calibrate procedure to %promote 
remedy for conventional phase correction procedures.
The calibration procedure allows phase correction to conveniently distinguish noise sign-symbols, eliminating the confusion between the noise-floor and normal signals. 
% The calibration compomises the coordinate discrepancy between complex-rotation, defined in a polar coordinate, and noise-floor, defined in a Cartesian coordinates, ultimately . 
% with the goal of unbiased white matter microstructure estimation.  
 % With an interesting finding that, the noise with different sign symbols is not treated approciately. 
%the phase correction procedures misleadingly generate the negative value, which is not caused by the noise-floor. 
%confuse the noise contributions in each complex coordinates; 
%and luckily found that a relative information can be used to fix this confusion.
%The calibration procedure is simple, easy to implement, and distinctly effective. 
Extensive experimental results, including those on both synthetic and real diffusion MRI data, demonstrate that calibrated procedures effectively help mitigate artifacts in phase-corrected DW images and FA maps, with improved accuracy on estimating FA in particular. 
% which is easy to implement and remedy the conventional procedures
% demonstrate that the calibrated procedures successfully eliminate artifacts in diffusion MR images and FA maps,  
% The experimental results suggest that patching the procedures of phase correction is necessary in generalizing the use of phase correction.
% Our work suggests that it should  necessarily take the filtering capacity into consideration in advancing phase correction technique.  % with the relative informance,

%The Diffusion-weighted (DW) magnitude is commonly Rician or non-central $\chi$ distributed, which is falsely elevated due to the noise-floor~\cite{eichner2015real,pizzolato2016noise}. To mitigate the noise-floor issue, phase correction extracts real-valued Gaussian-distributed diffusion MRI data,  eliminating biases of white-matter microstructure estimation (WM-MSE), \textit{e.g.}, fractional anisotropy (FA). Conventional phase correction methods resort to employing spatially-adaptive filters for improving perfromance~\cite{pizzolato2018automatic, liu2019mkf}. However, even MPPCA~\cite{veraart2016denoising}, renowned as spatial adaptivity, surprisingly introduces severe artifacts~\cite{liu2020gaussianization}. 

\section{Calibration of Phase Correction Procedures}
\label{sec2}
\rev{
     Diffusion MRI is inherently complex valued and low in SNRs. It could be corrupted by noise presented in both complex axes with changed signal sign-symbols.
     When reconstructing magnitude, the sum of square (SoS) operation simply flips all signal to positive. Thereby, negative contribution of magnitude turns to positive, which elevates magnitude and cannot be eliminated through averaging. 
     While noise-floor emerges under the case that SoS neglects noise sign-symbols, significant signal loss appears under the case that phase correction neglects the sign-symbols. 
     This issue has not been sufficiently studied for the long term seemingly because the negative contribution to magnitude cannot be analyzed in the commonly used polar coordinate.
     In this section, with a goal to unveil how phase correction inherently introduces artifacts to images, we utilize the complex polar coordinate system to comb through phase correction procedures and noise-floor (Section~\ref{sec21}), and accompanied with a generalized definition of noise-floor, we put forward a rule of thumbnail for \revvv{discerning} noise-floor accurately (Section~\ref{sec22}).
}
%%%we employ the complex polar coordinate system, which can be considered as a combination of Cartesian coordinate and polar coordinate, to cautiously analyze phase correction procedures and recharacterize the noise-floor. 

% Phase correction complex-rotates the true signal, instead of reconstruction magnitude, with the goal of avoiding noise-floor. However, it introduces artifacts to phase-corrected DW images, biasing the estimation of diffusion indices. In this section, we will illustrate the inherent problems in the procedures of phase correction and carlibrate them. How phase correction induces artifacts will be discussed in section~\ref{sec21}, while we will give detailed description of the carlibration of phase correction procedures in section~\ref{sec22}.

 % introduce a cooordinate system combining Cartesian coordinates with polar coordinate, the complex-rotation and noise-floor, both terminologies summarize processes that take place in different coordinate systems.  

\subsection{\rev{Indiscernibility of Noise Sign-Symbols \revvv{Results} in Signal Loss}} % Analyses of Phase Correction Procedures
\label{sec21}

%Phase correction complex-rotates \rev{and projects magnitude  to phase-corrected image without noise-floor}, utilizing the background phase that is estimated using a smoothing filter. %% Phase correction procedures can be analyzed in a complex polar coordinate system, which can be considered as a combination of polar coordinate and Cartesian coordinate. 

The complex-rotation employed by phase correction can be formulated in a polar coordinate: 
\begin{equation}
	I^{\text{PC}}(\bm{x}) = M(\bm{x}){\rm e}^{j\revv{ \Delta\varphi(\bm{x}) }}\,,
	\label{eqn:polar}
\end{equation}
where $\bm{x}$ denotes a voxel location and \revv{ $\Delta\varphi(\bm{x}) = \varphi(\bm{x}) - \varphi_{\text{BG}}(\bm{x})$} \revvv{is the rotation angle}. Given the noisy magnitude $M(\bm{x})$ and the phase $\varphi(\bm{x})$, the phase-corrected result $I^{\text{PC}}(\bm{x})$ is determined by the background phase $\varphi_{\text{BG}}(\bm{x})$,  
\begin{equation}
	\varphi_{\text{BG}}(\bm{x}) =  \arctan{\frac{f(I_{\text{i}}(\bm{x}))}{f(I_{\text{r}}(\bm{x}))}}\,,
	\label{eq2}
\end{equation}
where $f(\cdot)$ denotes a filtering operator on the real part $I_{\text{r}}(\bm{x})$ and the imaginary part $I_{\text{i}}(\bm{x})$ of DW signal,  respectively. 

At the same time, the complex-rotation can also be formulated in a Cartesian coordinate. %that composes of imaginary and real parts, 
In this case, the phase-corrected image $I^{\text{PC}}(\bm{x})$ can be written as
\begin{equation}%\label{eqn:cartesian}
   I^{\text{PC}}(\bm{x}) = I_{\text{r}}^{\text{PC}}(\bm{x})+j{I}_{\text{i}}^{\text{PC}}(\bm{x})\,.    
   \label{eqn:cartesian}
\end{equation}

Combining Eqn.\,\ref{eqn:polar} and Eqn.\,\ref{eqn:cartesian}, the phase-corrected real part $I_{\text{r}}^{\text{PC}}(\bm{x})$ and imaginary part $I_{\text{i}}^{\text{PC}}(\bm{x})$ are given by
%, a link between the two coodinate systems is the projection using phase,
\begin{equation}		
				\begin{cases}
				\begin{aligned}
					I_{\text{r}}^{\text{PC}}(\bm{x}) &= M(\bm{x})\cos(\revv{ \Delta\varphi(\bm{x}) })\,,\\
					I_{\text{i}}^{\text{PC}}(\bm{x}) &= M(\bm{x})\sin(\revv{ \Delta\varphi(\bm{x}) })\,.
				\end{aligned}
			\end{cases}
	\label{eq:e3}
\end{equation}
Please note that, by the complex-rotation, the imaginary part $I_{\text{i}}^{\text{PC}}(\bm{x})$ purely contains noise and will \revvv{be} discarded in phase correction, while the real part $I_{\text{r}}^{\text{PC}}(\bm{x})$ is used as the corrected magnitude corresponding to $M(x)$.

\begin{figure}[tb] 
	\centering
	%\subfigure[]{
	%\label{fig1} 
	%	\includegraphics[width=0.45\textwidth][../Fig/sub2]%{../Fig/sub2} 
	%}
	\subfigure{
	\label{fig1}
		\includegraphics[width=0.3\textwidth]{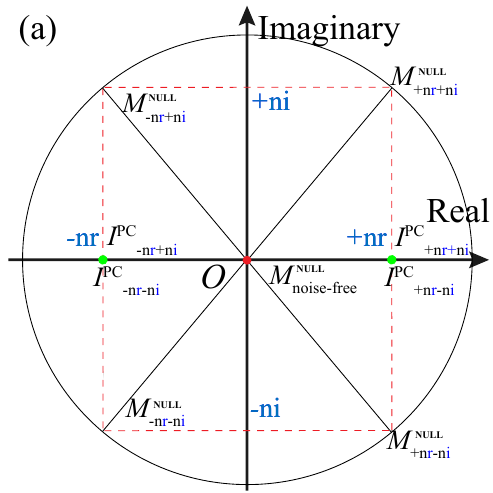}
	} 	%\\
    \hspace{10mm}
	\subfigure{
	\label{fig2}
		\includegraphics[width=0.3\textwidth]{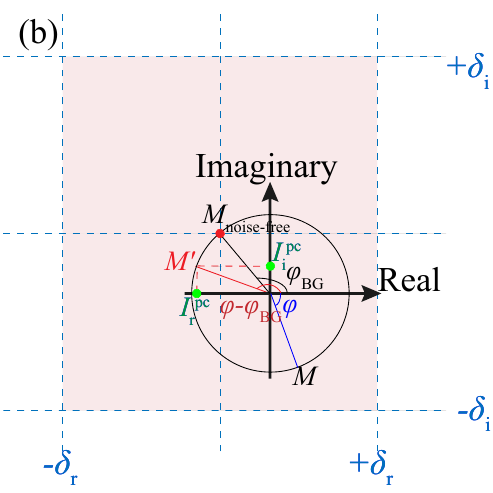} 
	}
	\\
	%\vspace{-5mm}
	%\subfigure{
	%	\includegraphics[width=0.45\textwidth]{../Fig/sub4}
	%}
	\caption{(a) The complex polar coordinate system, \revvv{with a noise-free complex voxel that is in absence of signal at the origin (\textit{i.e.},  $M^{\text{NULL}}_{\text{noise-free}} = 0$}), enables to analyze the negative contribution to magnitude.  % is employed to analyze the phase correction procedures. 
	The negative contribution \revvv{is derived from noise-floor}, denoted by  $I^{\text{PC}}_{-\text{nr}, - \text{ni}}$. (b) The noise-floor deflects noise-free voxel $M_{\text{noise-free}}$ to $M$ \rev{that} locates at the diagonal quadrant of the opposite side. \revv{$\delta_{\text{r}}$ and $\delta_{\text{i}}$ denote that the noise level in imaginary and real parts are the same.}} %24
	\label{fig_1}    
\end{figure}

%\begin{figure}[t] 
%	\centering
%	\includegraphics[width=0.7\textwidth]{../Fig/Figure0} 
%	\vspace{-8mm}
%	\caption{ The complex polar coordinate system \rev{enables to analyzing the negative contribution to magnitude. } % is employed to analyze the phase correction procedures. 
%	\rev{With a noise-free complex voxel that is in absence of signal at the origin, \textit{i.e.},  $M^{\text{NULL}}_{\text{noise-free}} = 0$,  the negative contribution of noise-floor is denoted by  $I^{\text{PC}}_{-\text{nr} - \text{ni}}$.} %  that is deviated by both real and imaginary noises. %, whose direction can be either negative or positive. 
%	% As anticipated, the positive and negative noises within an image region would be eliminated if they are averaged across each other. 
%	Conventional phase correction inherently \rev{makes} $I^{\text{PC}}_{-\text{nr} + \text{ni}}$ be falsely as small as $I^{\text{PC}}_{-\text{nr} - \text{ni}}$ \rev{, causing arbitrary signal loss, \textit{i.e.}, artifacts}.	
%   % Thus, the positive and negative noises should be treated differently, 
%     } %24
%	\label{fig1}       % Give a unique label	
%\end{figure}
%\todo{clarify the $M$, $M-{noisefree}$; emphaze the the most simple condition.}

To unveil how phase correction inherently introduces artifacts to images, we borrow the complex polar coordinate system, which is a mixture of the Cartesian and the Polar coordinate systems, and study \revvv{noise-floor in a simplified scenario} as shown in Fig.\,\ref{fig1}. 
We specifically exploit a noise-free complex voxel that is in absence of signal, with the magnitude $M^{\text{NULL}}_{\text{noise-free}} = 0$,  marked by the red dot at the origin, \textit{i.e.}, $Z = 0 + j0$. 
Let us suppose \rev{$M^{\text{NULL}}_{\text{noise-free}}$ is corrupted by four cases of complex noises, all of which have the same values for the real and the imaginary components (\textit{i.e.}, $\text{nr}= \text{ni}>0$) but with different sign-symbols (\textit{i.e.}, $\pm\text{nr}$ and $\pm\text{ni}$).} %Although $M_{-\text{nr}, -\text{ni}} = M_{+\text{nr}, +\text{ni}}=M_{-\text{nr}, +\text{ni}} = M_{+\text{nr}, -\text{ni}}$, 
\rev{Accordingly, this leads to four corrupted voxels with their magnitudes (denoted as $M_{-\text{nr}, -\text{ni}},M_{-\text{nr}, +\text{ni}}, M_{+\text{nr}, -\text{ni}}, \text{and}  M_{+\text{nr}, +\text{ni}}$ and located at different quadrants in the complex polar coordinate system). And their corresponding phase corrected magnitudes are denoted as $I^{\text{PC}}_{-\text{nr}, -\text{ni}}, I^{\text{PC}}_{-\text{nr}, +\text{ni}}, I^{\text{PC}}_{+\text{nr}, -\text{ni}}, \text{and}~     I^{\text{PC}}_{+\text{nr}, +\text{ni}}$, respectively, with their relationships shown in Fig.\,\ref{fig1}.} 
%\todo{Explain the four cases of noise. Emphaze the differece of two coordinates.}
% It is worth noting that noises in the real and imaginary parts are the same (\textit{i.e.}, $\text{nr} = \text{ni}$), if neglect the sign-symbol. 

Let us consider only $M_{-\text{nr}, -\text{ni}}$ (in the third quadrant) and $M_{+\text{nr}, +\text{ni}}$ (in the first quadrant) first.
When reconstructing $M_{-\text{nr}, -\text{ni}}$, the noises are typically flipped from negative to positive, resulting in noise-floor \revv{that} elevates magnitude in a postprocessing average step, due to $M_{-\text{nr}, -\text{ni}} + M_{+\text{nr}, +\text{ni}} > 0$. 
%\todo{More depiction on this process.}
For this case, the phase correction could \revv{avoid the noise-floor} since $I^{\text{PC}}_{-\text{nr}, -\text{ni}} + I^{\text{PC}}_{+\text{nr}, +\text{ni}} = 0$,  \rev{ assuming $\varphi^{\text{NULL}}_{\text{noise-free}}=0$}. % which successfully eliminates the noise-floor,

However, \revv{Fig.\,\ref{fig1} also demonstrates an inherent issue of phase correction procedures.} When considering $M_{-\text{nr}, -\text{ni}}$ (in the third quadrant) and $M_{-\text{nr}, +\text{ni}}$ (in the second quadrant), their phase corrected results become
$I^{\text{PC}}_{-\text{nr}, +\text{ni}} = I^{\text{PC}}_{-\text{nr}, -\text{ni}}$, which is quite counter-intuitive, because  
% The two cases of complex noises negatively contribute to the noise-floor should be different, $I^{\text{PC}}_{-\text{nr} + \text{ni}}$ is derived from positive imaginary noise while $I^{\text{PC}}_{-\text{nr}, -\text{ni}}$ is derived from negative imaginary noise.   causing arbitrary signal loss.
 %Fig.~\ref{fig1} indicates that phase correction cannot distinguish them since
in the context of noise-floor $I^{\text{PC}}_{-\text{nr}, +\text{ni}}$ is falsely as small as $I^{\text{PC}}_{-\text{nr}, -\text{ni}}$ after the correction. This may cause arbitrary signal loss and thus artifacts (such as black holes) in DW images and FA maps. 
%With this gusess, we implement strict analysis in the following section. }

% The formation of noise-floor is very sensitive to the sign-symbol of noise, however, we can find in  Fig.~\ref{fig1} that  phase correction  misleadingly results in $I^{\text{PC}}_{-\text{nr}, +\text{ni}} = I^{\text{PC}}_{-\text{nr}, -\text{ni}}$.
%the value of $I^{\text{PC}}_{-\text{nr} + \text{ni}}$ is falsely as low as that of $I^{\text{PC}}_{-\text{nr} - \text{ni}}$, and this under-estimation of $I^{\text{PC}}_{-\text{nr} + \text{ni}}$ \rev{incurs} the \rev{signal loss} (\textit{i.e.}, dark holes) to the phase-corrected images. 
%\todo{Why filter-based phase correction cannot handle the sign-symbol? phase correction procedures, specifically, the projection to one axis would neglect the noise information come from another axis. }
% and that is the reason why conventional phase correction falsely introduces dark holes.

\subsection{Noise-Floor Restatement and Phase Correction Calibration} % generalized definition
% Restatement redefinition.
\label{sec22}
%As demonstrated by above analyses,  phase correction %is %incapable of distinguishing the sign-symbols of noises, %which
%\revv{potentially confuses normal signals $I^{\text{PC}}_{-\text{nr}, +\text{ni}}$} with the true noise-floor $I^{\text{PC}}_{-\text{nr}, -\text{ni}}$. However, such definition is not strict. 
Inspired by the above observation, in this section, we first give a lemma to generalize the scenarios causing noise-floor,  then propose a theorem to unveil how to discern the noise-floor from normal signals, and ultimately develop a calibration remedy for the conventional phase correction. % , which is originally detected from the magnitude.   
\textit{Given a perfect filter}, % that is able to estimate noise-free $M_{\text{noise-free}}$, 
\revv{we can find that noise-floor occurs when the following condition is met. } %potentially  
%\rev{generalized from the polar coordinate to the complex polar-coordinates. } %, shown in Fig.~\ref{fig2} demonstrates the generalized definition of noise-floor. 
%%% previous statements are the simple case.
% demonstrates the voxels that should be noise-floor with high confidence. 
\rev{
    \begin{lemma}
    \revv{The noise-floor is caused by noises with different sign-symbols to the low-SNR complex signals, typically appearing when the complex noises deflect the noise-free signal to the diagonal quadrant on the opposite side in the complex polar coordinate system. % noises are not only change the magnitude, but also rotate the signals. 
    } 
	 \label{lm1}
    \end{lemma}
}
% \rev{The definition of noise-floor should be generalized from the polar coordinate to the complex polar-coordinates. }
%Thus, it should generalize of `negative' noise to the complex signals, and also the definition of noise-floor that is originally defined for magnitude signal to the complex signals.
% the noise-floor is noticed from magnitude, however, we need to discern it from the complex parts.  
% the difference between magnitude and complex signal, 
\rev{
	\revv{Lemma~\ref{lm1} can be pictorially illustrated by Fig.\,\ref{fig2}. 
	Compared with the magnitude $M^{\text{NULL}}_{\text{noise-free}}$ that is in absence of signal in Fig.\,\ref{fig1}, the real and imaginary signals of a low-SNR voxel $M_{\text{noise-free}}$ may have either positive or negative sign-symbols. 
	Thus, the concept of `negative' contribution to magnitude should be generalized accordingly. 
	From the given case in Fig.\,\ref{fig2} (more cases are in our supplementary materials), the sign-symbols of both signals are changed when super-imposed with noises. 
	Thus, the sum of square approach falsely accumulates the noise rather than the true signal because the inverted signal with different sign-symbol is purely noise. Even dramatically, this pure noise cannot be simply suppressed by averaging in postprocessing procedures, because its distribution has been changed to Rician or non-central $\chi$. }
	\begin{theorem}
		\label{t1}
			The noise-floor, which causes the noise-free signal $M_{\text{noise-free}}$ to deflect to the opposite diagonal quadrant, always has a complex-rotation angle %(\textit{i.e.}, the phase of $M^'$)  
			that is between $\pi/2$ and $3\pi/2$, and the resulting noise-floor-superimposed magnitude $M^'$ always locates at the second or the third quadrant, and vice versa. %, with a anglein rotating $M_{\text{noise-free}}$ to $M$
	\end{theorem}
}
\begin{proof}
	\textit{\textbf{Sufficiency}}: Because when rotating a complex signal to the opposite diagonal quadrant, the rotation should pass a full quadrant; we have $\revvv{\Delta\varphi(\bm{x}) > \pi/2}$, while  $\revv{ \Delta\varphi(\bm{x}) } < 3\pi/2$ is obvious. In addition,  since the $\revv{ \Delta\varphi(\bm{x}) }$ % should be the 
	is the phase of the complex-rotated magnitude $M^'$, %the latter therefore 
	\revvv{leading to $M^'$ always  locating} at the second or the third quadrant within the range $[\pi/2, 3\pi/2]$.
	%According to the definition of the noise-floor, a noisy singal $M$ with noise-floor must at the diagonal quadrant of the opposite of $M_{\text{noise-free}}$; thus, $\pi/2 < (\varphi(\bm{x}) - \varphi_{\text{BG}}(\bm{x})) < 3\pi/2$,  given a angle range $[0, 2\pi]$; 
	\textit{\textbf{Necessity}}: Given a noisy signal $M$ \revvv{at the opposite diagonal quadrant of $M_{\text{noise-free}}$}, %with $\pi/2 < \revv{ \Delta\varphi(\bm{x}) } < 3\pi/2$, 
	assume it is not noise-floor. This relationship reveals that \revvv{the sign-symbols of complex signal of $M_{\text{noise-free}}$ are totally changed.} Then according to Lemma~\ref{lm1}, $M$ is superimposed by noise-floor, which contradicts with the assumption. The proof completes.
	%thus the complex noises would not negatively contribute to the magnitude, however, the noisy signal  with . Hereby, the . \revv{it seems cause problems, the definition of the diagonal quadrant of the opposite side of $M_{\text{noise-free}}$ is not $\pi/2 < (\varphi(\bm{x}) - \varphi_{\text{BG}}(\bm{x})) < 3\pi/2$, it should be more strict. Or there needs a function to detect this characteristices.}
\end{proof}

Guided by Theorem 1, we propose a calibration procedure for phase correction as follows:
	\begin{equation}
		I^{\text{PC}}(\bm{x}) = M(\bm{x}){\rm e}^{j\varPsi(\revv{ \Delta\varphi(\bm{x}) })}\,.
	\label{eq5}
	\end{equation}
	\noindent $\varPsi(\cdot)$ \revv{denotes the operator to check whether $M$ is deviated by the complex noises to the opposite diagonal quadrant of $M_{\text{noise-free}}$. The calibration procedure is summarized in Algorithm~\ref{alg:A}. }
	% flips $\varphi(\bm{x}) - \varphi_{\text{BG}}(\bm{x})$ to the first quadrant such that $I^{\text{PC}}_{\text{i}}(\bm{x})$ is positive, and checks the noise sign-symbols according to theorem~\ref{t1} such that it only gives negative values to signals that is descerned as noise-floor. 
\begin{algorithm}
	\caption{Phase calibration for discerning noise-floor}
	\label{alg:A}
	\hspace*{0.02in} {\bf Input:} 
	$\varphi(\bm{x})$ and  $\varphi_{\text{BG}}(\bm{x})$ estimated by Eqn.\,\ref{eq2}\\
	\hspace*{0.02in} {\bf Output:} 
	$\Delta\varphi(\bm{x})$ to be used in Eqn.\,\ref{eq:e3} for phase correction
	\begin{algorithmic}[1]
		\REPEAT 		  %  \revv{ $\Delta\varphi(\bm{x}) = \varphi(\bm{x}) - \varphi_{\text{BG}}(\bm{x})$}	
		%\STATE %
		\STATE {\revvv{set $\Delta\varphi(\bm{x}) = \varphi(\bm{x}) - \varphi_{\text{BG}}(\bm{x})$;}}
		\STATE {if $\Delta\varphi(\bm{x})$ is in the $2\text{-nd}$ or $3\text{-rd}$ quadrant, then flip it to the  $1\text{-st}$ or $4\text{-th}$ quadrant, respectively, marked by $\Delta\varphi^{\text{f}}(\bm{x})$;}
		\STATE	{if $\Delta\varphi(\bm{x})$ deflects $M_{\text{noise-free}}$, estimated from filtering real and imaginary signals, to $M$ that locates at the opposite diagonal quadrant, then flip $\Delta\varphi^{\text{f}}(\bm{x})$ back to the $2\text{-nd}$ or $3\text{-rd}$ quadrant, marked by $\Delta\varphi^{\text{ff}}(\bm{x})$;}	
		\STATE {if $\Delta\varphi(\bm{x}) \neq \Delta\varphi^{\text{ff}}(\bm{x})$, then replace  $\Delta\varphi(\bm{x})$ by $\Delta\varphi^{\text{f}}(\bm{x})$;}
		%\STATE set $\Delta\varphi(\bm{x}) = \Delta\varphi(\bm{x})$ 
		\UNTIL{all voxels have been processed.} 
	\end{algorithmic}
\end{algorithm}

\section{Experimental Results}

\subsection{Datasets}

We employed both synthetic and real data to demonstrate the effectiveness of the calibrated procedures for phase correction.
We synthesized the background phase according to \cite{pizzolato2016noise}, and generated synthetic DW data using Phantom$\alpha$s~\cite{caruyer2014phantomas}, with the same gradient vectors of the real DW data described as follows. 
With the noise-free DW data, we added spatially-varying Gaussian noises to the real and imaginary parts, respectively\rev{~\cite{chen2019denoising}}. 
The real DW data were acquired using a SIEMENS $3$T Magnetom Prisma MR scanner with following protocol, TR =$2500\,\text{ms}$, TE=$89\,\text{ms}$, FoV=$210\times 210\,\text{mm}^2$, matrix size=$140\times140$, and $b=750,1500,3000\,\text{s/mm}^2$ with a total of $64$ diffusion directions.

\subsection{Experimental Settings} 
\label{subsec:S31}
We chose the optimal filtering parameters based on the synthetic data and applying them to processing the real DW signal. To verify the effectiveness of our proposed calibration method, we compared phase correction performance before and after \rev{the} calibration. The parameters used in the phase correction are set as follows. For TV denoising (denoted as TV), the regularization parameter $\lambda$ was set as $2$, and the iteration limit was set as $10$. For Curvature Filtering (denoted as CF), the iteration limit used by Gaussian curvature filter was set as $10$. For MPPCA denoising (denoted as MPPCA), the block size was set as $5\times5\times5$.

% \item \textbf{Weighted TV denoising:} To deal with spatially-varying noise, \cite{pizzolato2018automatic} extended TV denoising by setting the regularization parameter spatially-adaptively according to the local noise level. To estimate the noise level, a local window of size $15$ voxels was used \cite{pizzolato2018automatic}. Other parameters were chosen to be identical to the TV method described above. 		

%\item \textbf{Curvature filtering (CF)~\cite{gong2017curvature}:} 
%We set Gaussian curvature filter with a iteration limit of $10$.

%\item \textbf{MPPCA denoising:} We set a block size of $5\times5\times5$. 

%\end{itemize} 
% We first employed the baseline methods to smooth the real and imaginary images with the same set of parameters. We then estimated the background phase using the smoothed images.

\subsection{Evaluation Criteria}
The calibrated phase correction is expected to mitigate artifacts in DW images and FA maps. To verify this, on the synthetic data, since the ground-truth is known, we could quantitatively calculate the mean absolute error (MAE, \textit{i.e.}, the voxel-wise absolute error averaged across a volume) between the generated and the ground-truth DW images and the mean error (ME, \textit{i.e.}, average voxel-wise error) within the white-matter regions between the generated and the ground-truth FA maps. On the real data, since the ground-truth is unavailable, we could only qualitatively evaluate the performance based on the criterion that a good phase correction is expected to lead to enhanced contrast in DW images and increased FA values in the white-matter regions, without introducing artefacts.

\subsection{Results And Discussions}
We compared 
\begin{inparaenum}[(i)]
\item the phase correction performance before and after calibration, and
\item the performance of  three different filters under the same experimental setting,  % our proposed method with that of another
\end{inparaenum}
including CF~\cite{gong2017curvature}, TV~\cite{rudin1992nonlinear}, and MPPCA~\cite{veraart2016denoising}. Their counterparts using the proposed calibration procedure are denoted as ``CF-new", ``TV-new", and ``MPPCA-new", respectively.

\begin{figure} [t]
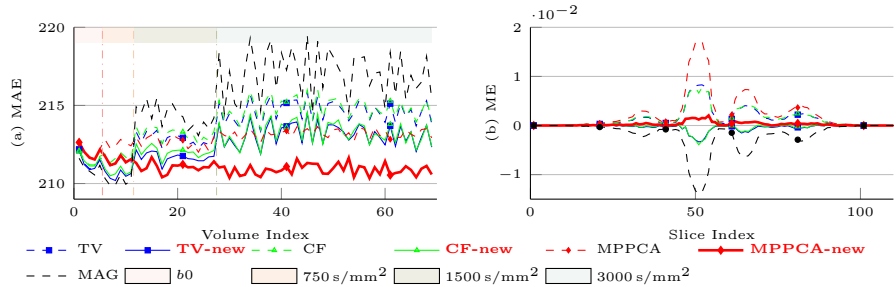

	\centering
	\subfigure{
		\label{fig:mae}
		\subimport{Fig/}{mae.tex} 
	}
	\hspace{1mm}
	\subfigure{
		\label{fig:me}
		\subimport{Fig/}{me.tex}
	}
    \\
\definecolor{mycolor1}{rgb}{0.98820,0.61570,0.60390}%
\definecolor{mycolor2}{rgb}{0.97650,0.80390,0.67840}%
\definecolor{mycolor3}{rgb}{0.78430,0.78430,0.66270}%
\definecolor{mycolor4}{rgb}{0.51370,0.68630,0.60780}%
\vspace{-0.4cm}
	\ref{legend:CompMae}
	\\
	%\vspace{-0.4cm}
	\caption{Quantitative comparison on the synthetic data. (a) MAE on each DW image (volume), and (b) ME of each slice on FA maps.} % \textbf{}
	\label{fig:curves}       % Give a unique label
\end{figure}

\noindent\textbf{Results on synthetic data}~~Fig.\,\ref{fig:curves} shows the quantitative results on synthetic data. On the left shows the MAE of DW images, and on the right shows the ME of FA maps. MAG (the black dashed line) denotes the magnitude of DW signals without phase correction. As can be seen, it produces much higher MAE on DW images, as well as higher ME on FA maps, than those phase correction based methods. Meanwhile, comparing phase correction with (indicated by solid lines) and without (indicated by dashed lines) the proposed calibration procedure, it can be seen that by using calibration, the produced MAE of DW image and ME of FA maps could be significantly reduced. Among all methods, MPPCA-new (\textit{i.e.}, MPPCA with calibration) achieves the best performance with the lowest MAE and ME values.

%the  phase corrected DW images achieves smaller MAE than MAG, while calibrated procedures are smaller than the conventional procedures. MPPCA\_new achieves the best performance in the most of the volumes with  large $b$-values.  The better phase-corrected DW images assure improvement in computing FA, Fig.~\ref{fig:me} shows that MPPCA\_new also achieves the best performance. We can also find that  the MAG  FA is significantly decreased by the noise-floor.    

\usetikzlibrary{backgrounds,arrows}
\usetikzlibrary{arrows,shapes,chains}
\usetikzlibrary{automata, chains, decorations.pathmorphing, positioning}
\usetikzlibrary{shapes.geometric, arrows,shapes.multipart}
\usetikzlibrary{fadings}

\let\iwidth\relax
\let\iheight\relax
\let\rowspace\relax
\newlength{\iwidth}
\newlength{\iheight}
\newlength{\rowspace}
\setlength{\iwidth}{0.1\textwidth}
\setlength{\iheight}{1.0\iwidth}
\setlength{\rowspace}{0pt}

\setlength{\tabcolsep}{1.5pt}

\newlength{\colorbarlength}
\setlength{\colorbarlength}{0.4\iheight}

\newcommand{\columnText}[1]{
	\begin{tikzpicture}[baseline,trim right=5pt]
		\node[rotate=90] at (0.05\iwidth, 0.5\iheight) {\tiny{#1}};
	\end{tikzpicture}
} 

\newlength{\gbarlength}
\newcommand\gBar[2]{		
	\begin{tikzpicture} %[baseline=(current bounding box.south)]
		\node (bar) {\pgfuseshading{g1}}; 
		\node [below=0mm of bar] {\scriptsize{#1}};
		\node [above=-0.3mm of bar] {\scriptsize{#2}};
	\end{tikzpicture}
} 	

\newcommand\colorBar[2]{		
	\begin{tikzpicture} %[baseline=(current bounding box.south)]
		\node (bar) {\pgfuseshading{spectrum1}}; 
		\node [below=0mm of bar] {\scriptsize{#1}};
		\node [above=-0.3mm of bar] {\scriptsize{#2}};
	\end{tikzpicture}
}

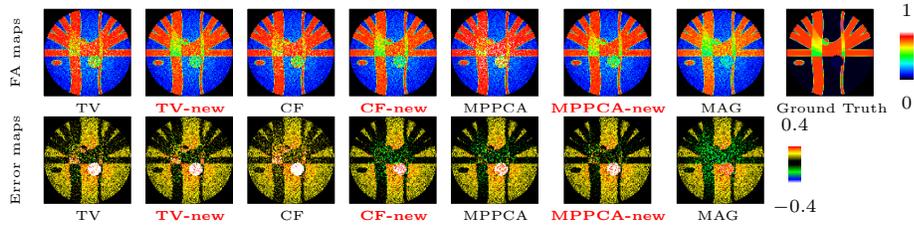
\begin{figure}[t] 
\centering
    \begin{tikzpicture}
             \pgfdeclareverticalshading{spectrum1}{.16cm}{
             	rgb(0)=(0,0,0); 
             	rgb(0.13\colorbarlength)=(0,0,1); 
             	rgb(0.43\colorbarlength)=(0,1,1); 
             	rgb(0.65\colorbarlength)=(0,1,0); 
             	rgb(0.86\colorbarlength)=(1,1,0); 
             	rgb(1.17\colorbarlength)=(1,0,0); 
             	rgb(1.30\colorbarlength)=(1,1,1)
             }
        \node(dum) {\tikzset{
	/tpic/options/.style = {
		width=\iwidth,
		height=\iheight,
		scale=1.05,
		xloc=0.5,
		yloc=0.5,
		angle=0,
		color=white, 
		label={}
	}
}
	\begin{tabular}{ccccccccc}
		\columnText{FA maps} &
		\tpic[options]{sim/fa_tv} & \tpic[options]{sim/fa_tv_500} & \tpic[options]{sim/fa_cf} & \tpic[options]{sim/fa_cf_500} &
		\tpic[options]{sim/fa_mppca}   &  \tpic[options]{sim/fa_mppca_500} & \tpic[options]{sim/fa_mag_500} &
		\tpic[options]{sim/fa_MAG_nf} \vspace{-2mm}  
		\\					
		[0.6\rowspace] & \tiny{TV} & \tiny{\textbf{\textcolor{red}{TV-new}}} & \tiny{CF} & \tiny{\textbf{\textcolor{red}{CF-new}}} & \tiny{MPPCA}  & \tiny{\textcolor{red}{\textbf{MPPCA-new}}} & \tiny{MAG}  &  \tiny{Ground Truth} 
		\vspace{-2mm}  
		\\
		\vspace{-2mm} 
		\columnText{Error maps} &
		\tpic[options]{res/res_tv_real} & \tpic[options]{res/res_cf_real} &
		\tpic[options]{res/res_mppca_real}  & \tpic[options]{res/res_tv_real_new} & \tpic[options]{res/res_cf_real_new} & \tpic[options]{res/res_mppca_real_new} & \tpic[options]{res/res_mag_final} 		&
		\\
			[0.6\rowspace] & \tiny{TV} & \tiny{\textbf{\textcolor{red}{TV-new}}} & \tiny{CF} & \tiny{\textbf{\textcolor{red}{CF-new}}} & \tiny{MPPCA}  & \tiny{\textcolor{red}{\textbf{MPPCA-new}}} & \tiny{MAG} 
		\vspace{-2mm} 
		%\\
		%[0.6\rowspace] & \scriptsize{TV} & \scriptsize{CF} & \scriptsize{MPPCA} & \scriptsize{TV new} & \scriptsize{CF new} & \scriptsize{MPPCA new} & \scriptsize{MAG} &
	\end{tabular}

};
        \node(cb) [above right = -1.8cm and -0.3cm of dum] {\colorBar{${0}$}{${1}$}};
             \pgfdeclareverticalshading{spectrum1}{.16cm}{ 
                	rgb(0)=(1,1,1);
                	rgb(0.1\colorbarlength)=(0,0,1);
                	rgb(0.3\colorbarlength)=(0,1,0);                	
                	rgb(0.5\colorbarlength)=(0,0,0);  
                	rgb(0.55\colorbarlength)=(0,0,0);	
                	rgb(0.6\colorbarlength)=(0,0,0);                	
                	rgb(0.8\colorbarlength)=(1,1,0);
                	rgb(1.0\colorbarlength)=(1,0,0) %0.15               	 
                }
        \node(db) [below right = -1.6cm and -2cm of dum] {\colorBar{${-0.4}$}{${0.4}$}};
    \end{tikzpicture}
    \\
    %\vspace{-0.4cm}
    \caption{ Visual comparison on synthetic data. Top: FA maps; Bottom: Error maps.}%purple
	\label{fig3}
\end{figure}

% the renewed phase correction procedures even reduce the FA errors. 
Fig.\,\ref{fig3} provides a visual comparison of the phase-corrected FA maps derived from synthetic data. As can be seen, phase correction brings increased FA values in  white-matter regions when compared with the magnitude MAG without correction. It is also found that conventional phase correction significantly introduces artifacts that excessively enlarge FA. 
This phenomenon is double confirmed by the error maps shown in the second row, where the calibrated phase corrections ``FA-new", ``CF-new", and ``MPPCA-new" yield less errors as indicated by more black regions in the error maps, indicating an improved accuracy on estimating FA, especially for ``MPPCA-new".

\setlength{\gbarlength}{1.4\iheight}

	\let\iwidth\relax
	\let\iheight\relax
	\let\rowspace\relax
	\newlength{\iwidth}
	\newlength{\iheight}
	\newlength{\rowspace}
	\setlength{\iwidth}{1.4cm}
	\setlength{\iheight}{1.0\iwidth}
	\setlength{\rowspace}{-6pt}
	\setlength{\tabcolsep}{0pt}
	
\tikzset{
	/tpic/options/.style = {
		width=\iwidth,
		height=\iheight,
		scale=1.05,
		xloc=0.5,
		yloc=0.5,
		angle=0,
		color=white,
		label={}
	} 
}
   
\pgfdeclareverticalshading{g1}{.20cm}{ %0.05\colorbarlength
    	rgb(0)=(0,0,0); 
    	rgb(1.30\gbarlength)=(1,1,1)
}    
    
\begin{figure}[t]
\centering 	
		\begin{tikzpicture}
		\node(dum) {\tikzset{
		/tpic/options/.style = {
			width =\iwidth,
			height=\iheight,
			scale=1.05,
			xloc=0.5,
			yloc=0.5,
			angle=0,
			color=white,
			label={}
		}
}
\begin{tabular}{ccccccc} 
		%\columnText{$0\,\text{s/mm}^2$}     \tpic[options]{TV}   &  \tpic[options]{TV_real_noise}   &  \tpic[options]{CF}  &  \tpic[options]{CF_real_noise}  & \tpic[options]{MPPCA_real}  & \tpic[options]{MPPCA_real_noise}  &  \tpic[options]{data}  \vspace{-4mm}  \\
		%\columnText{$750\,\text{s/mm}^2$}   \tpic[options]{TV2}  &  \tpic[options]{TV_real_noise2}  &  \tpic[options]{CF2} &  \tpic[options]{CF_real_noise2} & \tpic[options]{MPPCA_real2} & \tpic[options]{MPPCA_real_noise2} &  \tpic[options]{data2} \vspace{-4mm}  \\ 	
		   %\\
		\tpic[options]{TV4}  &  \tpic[options]{TV_real_noise4}  &  \tpic[options]{CF4} &  \tpic[options]{CF_real_noise4} & \tpic[options]{MPPCA4}      & \tpic[options]{MPPCA_real_noise4} &  \tpic[options]{data4} 
		\vspace{-2mm} 
		\\
		\tpic[options]{tv_r}  &  \tpic[options]{tv_n_r}  &  \tpic[options]{cf_r} &  \tpic[options]{cf_n_r} & \tpic[options]{mppca_r} & \tpic[options]{mppca_n_r} &  \tpic[options]{data_r}  
		\vspace{-2mm}
		\\  
		\tiny{TV} &  \tiny{\textbf{\textcolor{red}{TV-new}}}             &  \tiny{CF}     &  \tiny{\textbf{\textcolor{red}{CF-new}}}            & \tiny{MPPCA}          & \tiny{\textbf{\textcolor{red}{MPPCA-new}}}            &  \tiny{MAG}
\end{tabular}

%\end{document}
};
		\node(cb) [right= -0.5cm of dum] {\gBar{$-20$}{$200$}};
		\end{tikzpicture}
   % \vspace{-1.5\baselineskip} 
   \\
   %\vspace{-5mm} 
    \caption{Phase-corrected DW images ($b=3000\,\text{s/mm}^2$). Red/yellow arrows denote dark holes introduced/mitigated  by conventional/calibrated phase correction.}
	\label{fig4}
\end{figure}
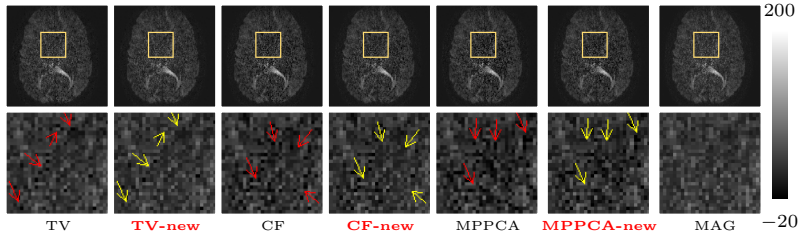

\noindent\textbf{Results on real data}~~Due to lack of ground-truth, qualitative analysis was conducted to investigate the proposed calibrated phase correction on real data.  Fig.\,\ref{fig4} demonstrates phase-corrected DW images with enhanced contrast when compared to MAG that is without correction. 
In addition, it is found that calibrated procedures yield much fewer dark holes 
at image regions as indicated by the red arrows \revvv{where intensities are coherently higher than surrounding regions}. % and the intensities produced by them are coherently higher than those of the conventional ones. 
%In addtion, we the calibrated phase correction procedures introduce significantly fewer dark holes to the phase-corrected real images, as shown in
As reported by \cite{liu2020gaussianization}, dark \revvv{holes are also reflected by artifact in FA maps shown} in Fig.\,\ref{fig5}. 
The calibrated phase correction successfully \rev{mitigates} dark holes in DW images as shown in Fig.\,\ref{fig4} and consequently results in clean FA maps as shown in Fig.\,\ref{fig5}. 
Moreover, we can also find that the corpus callosum region shows enlarged FA values, which is consistent with the results documented in  \cite{eichner2015real}. 
\usetikzlibrary{backgrounds,arrows}
\usetikzlibrary{arrows,shapes,chains}
\usetikzlibrary{automata, chains, decorations.pathmorphing, positioning}
\usetikzlibrary{shapes.geometric, arrows,shapes.multipart}
\usetikzlibrary{fadings}

\let\iwidth\relax
\let\iheight\relax
\let\rowspace\relax
\newlength{\iwidth}
\newlength{\iheight}
\newlength{\rowspace}
\setlength{\iwidth}{0.11\textwidth}
\setlength{\iheight}{1.0\iwidth}
\setlength{\rowspace}{0pt}

\tikzset{
	/tpic/options/.style = {
		width=\iwidth,
		height=\iheight,
		scale=1.05,
		xloc=0.5,
		yloc=0.5,
		angle=0,
		color=white,
		label={}
	}
}

\setlength{\tabcolsep}{1.5pt}

\setlength{\colorbarlength}{0.4\iheight}
\pgfdeclareverticalshading{spectrum1}{.20cm}{ %0.05\colorbarlength
	rgb(0)=(0,0,0); 
	rgb(0.13\colorbarlength)=(0,0,1); 
	rgb(0.43\colorbarlength)=(0,1,1); 
	rgb(0.65\colorbarlength)=(0,1,0); 
	rgb(0.86\colorbarlength)=(1,1,0); 
	rgb(1.17\colorbarlength)=(1,0,0); 
	rgb(1.30\colorbarlength)=(1,1,1)
}

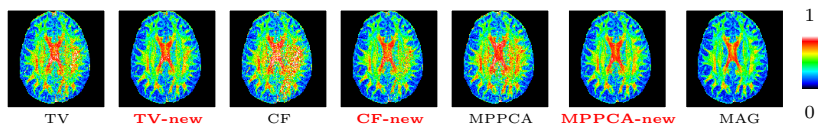
\begin{figure}[t] 
\centering
    	\begin{tikzpicture}
    	    \node(dum) {\tikzset{
	/tpic/options/.style = {
		width=\iwidth,
		height=\iheight,
		scale=1.05,
		xloc=0.5,
		yloc=0.5,
		angle=0,
		color=white,
		label={}
	} 
}
	\begin{tabular}{cccccccc}
		\tpic[options]{real/fa_TV} & \tpic[options]{real/fa_TV_1} &\tpic[options]{real/fa_CF} & \tpic[options]{real/fa_CF_1} & 		\tpic[options]{real/fa_MPPCA} &   \tpic[options]{real/fa_MPPCA_1} &		\tpic[options]{real/fa_MAG}  
	\vspace{-2mm}
		\\		
	 \tiny{TV} & \tiny{\textbf{\textcolor{red}{TV-new}}} & \tiny{CF} & \tiny{\textbf{\textcolor{red}{CF-new}}} &  \tiny{MPPCA} & \tiny{\textbf{\textcolor{red}{MPPCA-new}}} & \tiny{MAG} 
 	 	   % real/fa_MAG3
	%	\\
		% [0.6\rowspace] & \scriptsize{TV} & \scriptsize{CF} & \scriptsize{MPPCA}  
	\end{tabular}
 
};
    	    \node(cb)  [ above right = -1.9cm and -0.3 cm of dum]  {\colorBar{${0}$}{${1}$}};
    	    %\node(cb1) [below right = -2.3cm and -2.2 cm of dum]  {\colorBar{${0}$}{${1}$}};
        \end{tikzpicture} \\
        %\vspace{-5mm}
    \caption{ FA maps obtained by phase correction without and with the calibration. Artifacts are significantly mitigated for all three filtering methods, while the FA values in corpus callosum are increased properly, after using calibration. }
	\label{fig5}
\end{figure}

In sum, our results sufficiently demonstrate that conventional phase correction is also the cause of introducing artifacts. 
With the proposed simple calibration procedure, aforementioned artifacts could be successfully mitigated in both DW images and FA maps. % is proposed and  s
% An interesting phenomenon attracts our attention, the artifacts in both DW images and FA maps locate at the central regions where noise level is larger than peripheral. Because large noise level decreases SNR as well so that has more change to be confused with noise-floor, such a rule also indicate the calibrated procedures succefully eliminate such artifacts.
 %, which results further support that such a claim in reverse. where the artifacts introduced by conventional phase correction are effectively eliminated.  peripheric 

\section{Conclusions}
Recent efforts in mitigating artifacts in DW images lie in adapting filters to the spatial variability of noise. % introduced by phase correction 
In this paper, for the first time, we propose that even a perfect filter is insufficient for phase correction. %such an approach is just a half-measure. 
By combing through the phase correction procedures in the complex polar coordinate system, we find that they are incapable of distinguishing sign-symbols of noise and  confuse the noise-floor with other normal signals. %like the way reconstructing magnitude. 
To bridge this gap, we propose a theorem and tactically develop a calibration procedure as the remedy, which is conceptually simple but distinctively effective as demonstrated by our experimental results.

\bibliographystyle{llncs_splncs}
\bibliography{references}

\end{document}

\title{Supplementary Materials} 
 	% Does Perfect Filtering Contributs to Perfect Phase Correction?
 	
 	% % For anonymous submission
 	\authorrunning{Anonymous\\Anonymous\\Anonymous}
 	\author{Is Perfect Filtering Enough Leading to Perfect Phase Correction?}
 	\institute{Paper \#471}
 	\titlerunning{Is Perfect Filtering Enough?}
 	
 	%\authorrunning{F.~Liu, L.~Zhou, J.~Yang, X.~He, J.~Feng, D.~Shen}   % abbreviated author list (for running head)
 	
 	%\author{Feihong~Liu\inst{1,2} \and Luping~Zhou\inst{3}  \and Junwei~Yang\inst{4,2} \and Xiaowei~He\inst{1,5} \and Jun~Feng\inst{1,5}~\Letter \and   Dinggang~Shen\inst{2,6,7}~\Letter}
 	%\institute{  School of Information Science and Technology,       Northwest University,         Xi'an,      China            \and	  % 1
 	%	         School of Biomedical Engineering,                   ShanghaiTech University,      Shanghai,   China          	\and      % 2
 	%             School of Electrical and Information Engineering,   University of Sydney,         Sydney,     Australia        \and      % 3 
 	%             Department of Computer Science and Technology,      University of Cambridge,      Cambridge,  United Kingdom   \and      % 4
 	%	         State-Province Joint Engineering and Research Center of Advanced Networking and Intelligent Information Services, School of Information Science and Technology, Northwest University, Xi’an, China                                                                             \and      % 5
 	%	         Shanghai United Imaging Intelligence Co., Ltd., Shanghai, China                                                \and      % 6	                
 	%            Department of Artificial Intelligence, Korea University, Seoul, Republic of Korea \\                                     % 7
 	%             \email{fengjun@nwu.edu.cn},~\email{dgshen@shanghaitech.edu.cn}
 	%}

 	\maketitle

 	\begin{figure}[h] 
 		\centering
 		\subfigure{
 			\includegraphics[width=0.45\textwidth]{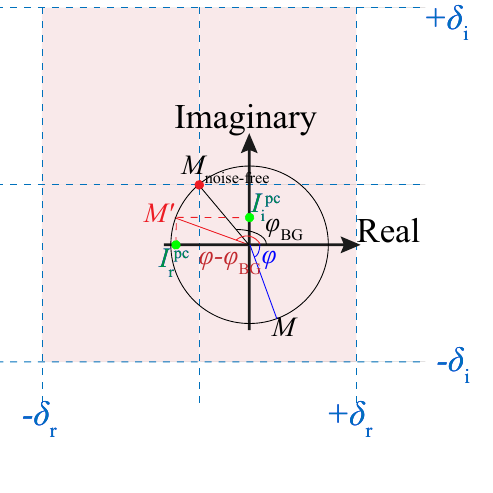} 
 		}
 		\subfigure{
 			\includegraphics[width=0.45\textwidth]{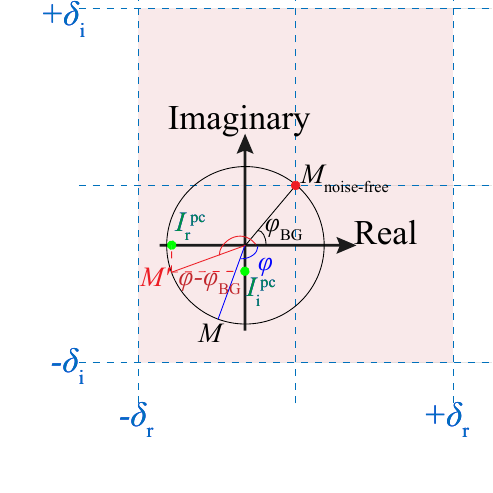}
 		} 
 		\vspace{-10mm}
 		\\
 		\subfigure{
 			\includegraphics[width=0.45\textwidth]{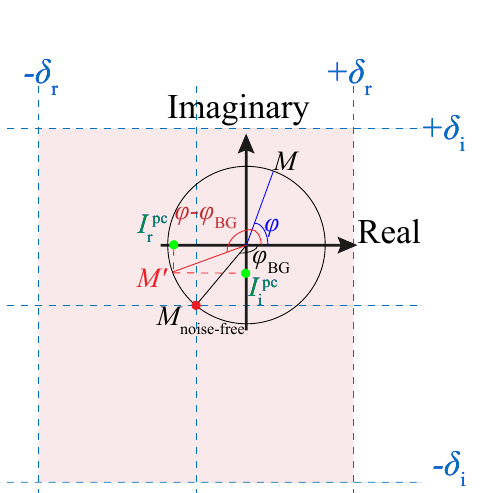} 
 		}
 		\subfigure{
 			\includegraphics[width=0.45\textwidth]{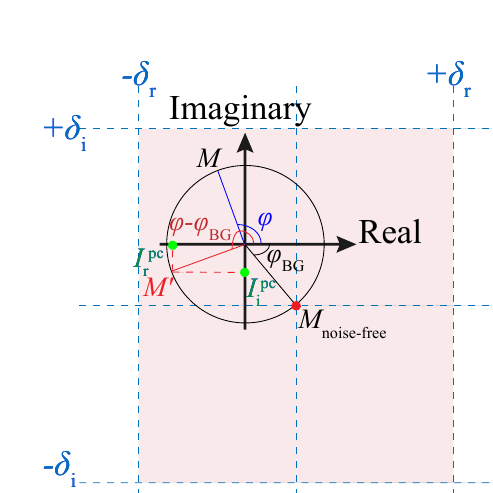}
 		}
 		\caption{The noise-floor deflects noise-free signal $M_{\text{noise-free}}$ to $M$ {that} locates at the diagonal quadrant of the opposite side. $\delta_{\text{r}}$ and $\delta_{\text{i}}$ denote that the noise level in imaginary and real parts are the same. } %24
 		\label{fig2}    
 	\end{figure}
 	
 	% \rev{The definition of noise-floor should be generalized from the polar coordinate to the complex polar-coordinates. }
 	%Thus, it should generalize of `negative' noise to the complex signals, and also the definition of noise-floor that is originally defined for magnitude signal to the complex signals.
 	% the noise-floor is noticed from magnitude, however, we need to discern it from the complex parts.  
 	% the difference between magnitude and complex signal, 

 From Fig.\,\ref{fig2}, the sign-symbols of both signals are changed when super-imposed with noises. 
 Thus, the sum of square approach falsely accumulates the noise rather than the true signal because the inverted signal with different sign-symbol is purely noise. Even dramatically, this pure noise cannot be simply suppressed by averaging in postprocessing procedures, because its distribution has been changed to Rician or non-central $\chi$.

% 	\input{../Fig/pc_real2/pc_realsb.tex}
 	
% 	Fig.~\ref{fig4} demonstrates phase-corrected DW images with enhanced contrast when compared to MAG. 
% 	In addition, we can find that calibrated \rev{procedures} yeild much fewer dark holes at image regions where intensities are coherently high than the conventional one. %In addtion, we the calibrated phase correction procedures introduce significantly fewer dark holes to the phase-corrected real images, as shown in